%% file: conference_101719.tex
\documentclass[conference, 10pt, letterpaper]{ieeeconf}
\IEEEoverridecommandlockouts
\usepackage{cite}
\usepackage{amsmath,amssymb,amsfonts,bm}
\usepackage{graphicx}
\usepackage{textcomp}

\usepackage{times}
\usepackage{epsfig}

\usepackage{hyperref}
\usepackage{url}

\usepackage{tabularx}
\usepackage{caption}
\usepackage{float}
\usepackage{makecell}
\usepackage{multirow, booktabs}
\usepackage{xcolor,colortbl}

\usepackage{subcaption}
\definecolor{Gray}{gray}{0.9}
\usepackage{xspace}
\def\baselinestretch{0.976}

\newcounter{para}

\DeclareRobustCommand\onedot{\@\xspace}

\def\eg{\emph{e.g}\onedot} 
\def\ie{\emph{i.e}\onedot}

\def\wrt{w.r.t\onedot}

\input{math_commands.tex}

\def\BibTeX{{\rm B\kern-.05em{\sc i\kern-.025em b}\kern-.08em
    T\kern-.1667em\lower.7ex\hbox{E}\kern-.125emX}}
\begin{document}

\title{Deep Bayesian ICP Covariance Estimation}

\author{Andrea De Maio$^{1}$ and Simon Lacroix$^{1}$
\thanks{$^{1}$LAAS-CNRS, Universit\'e de Toulouse, 7 Avenue du Colonel Roche, 31031, Toulouse, France,
\texttt{E-mail: andrea[dot]de[dash]maio[at]laas[dot]fr}%
}}

\maketitle

\begin{abstract}
Covariance estimation for the Iterative Closest Point (ICP) point cloud registration algorithm is essential for state estimation and sensor fusion purposes.
We argue that a major source of error for ICP is in the input data itself, from the sensor noise to the scene geometry. Benefiting from recent developments in deep learning for point clouds, we propose a data-driven approach to learn an error model for ICP. We estimate covariances modeling data-dependent heteroscedastic aleatoric uncertainty, and epistemic uncertainty using a variational Bayesian approach.
The system eva\-lua\-tion is performed on LiDAR odometry on different datasets, highlighting good results in comparison to the state of the art.
\end{abstract}

\section{Introduction}

With the steady increase in 3D sensors availability, point clouds have seen a rapid diffusion and adoption. 
Their use benefits from the 3D information that they readily express, which allows to easily retrieve geometric properties and yield the possibility to extract  complex geometric features to classify object shapes, perform scene segmentation, or assess complex situations.
Point clouds are also used to solve motion estimation and self-localization problems in robotics.
Scan registration algorithms provide an estimate of the 3D transformation between the positions at which two point clouds are acquired.
Such processes are at the heart of LiDAR-based odometry \cite{zhang2014loam} and SLAM frameworks \cite{mendes2016icp}.
Ori\-gi\-nal\-ly introduced in \cite{BESL-PAMI-1992}, the Iterative Closest Point (ICP) algorithm, in its many variants, gradually became the standard approach to the point-cloud registration problem \cite{pomerleau2015review}. 

\begin{figure}[]
\centering
\captionsetup{width=\linewidth}
\begin{subfigure}{\columnwidth}
\includegraphics[trim={0 0 0 1.2cm}, clip, width=\columnwidth]{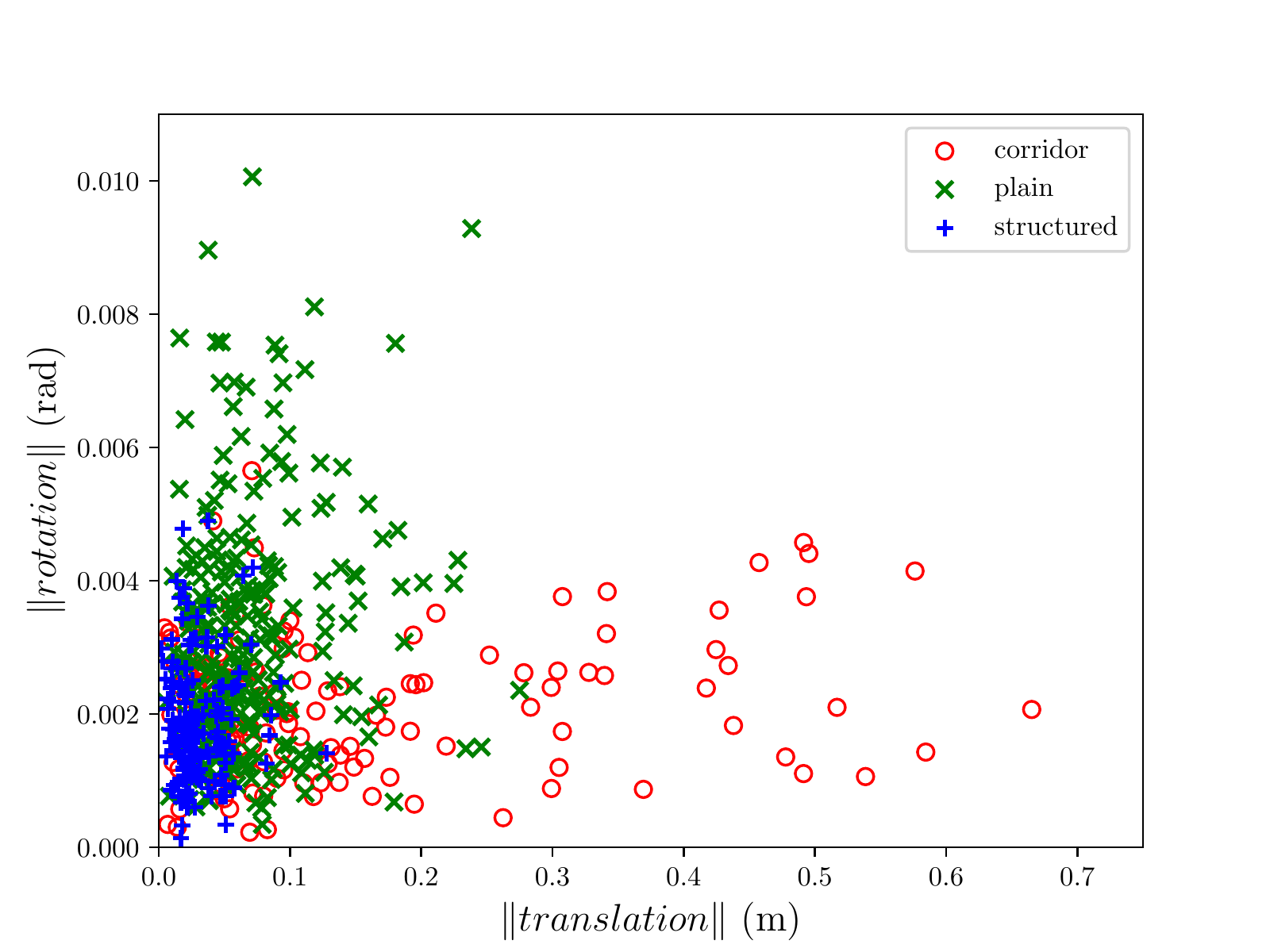}
\end{subfigure}
\\[0.2cm]
\begin{subfigure}{0.3\columnwidth}
\includegraphics[trim={2cm 0 13cm 0},clip,width=\linewidth]{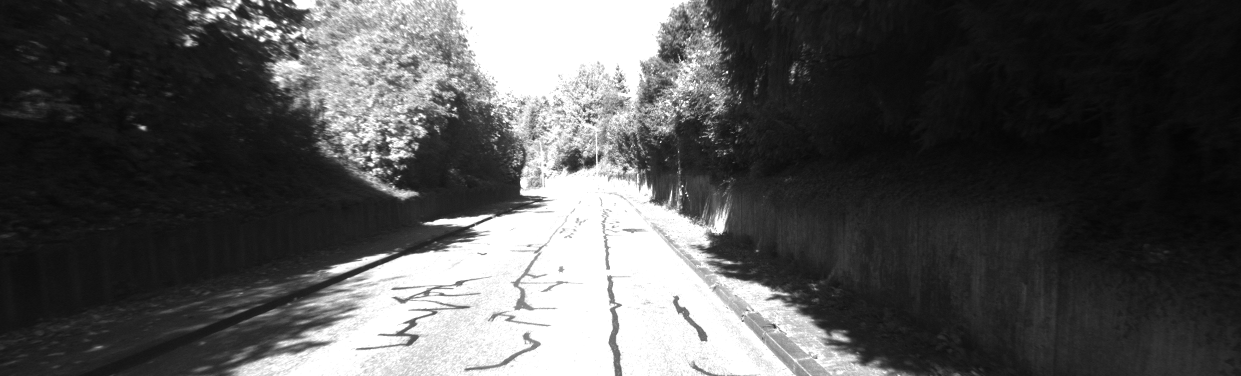}
\end{subfigure}
\begin{subfigure}{0.3\columnwidth}
\includegraphics[trim={7.5cm 0 7.5cm 0},clip, width=\linewidth]{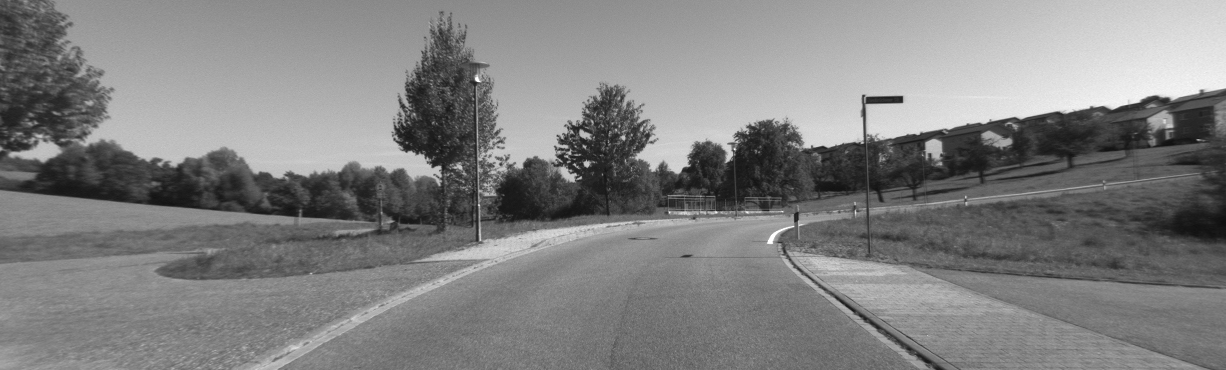}
\end{subfigure}
\begin{subfigure}{0.3\columnwidth}
\includegraphics[trim={5cm 0 10cm 0},clip, width=\linewidth]{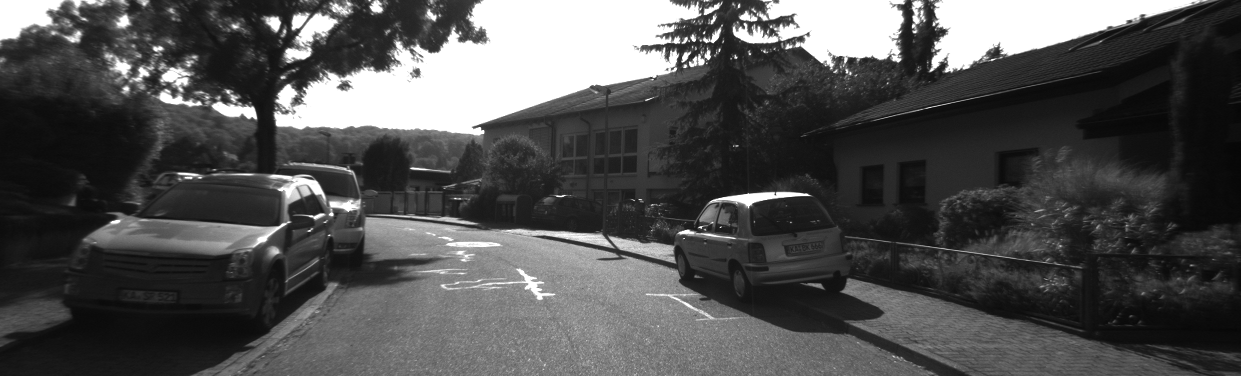}
\end{subfigure}
  \caption{Distribution of errors on a series of ICP estimates computed in three different kind of environments: corridor-like (left), wide plain (center), structured (right).}
\label{fig:first_page}
\end{figure}

In order to qualify the result of ICP, and especially to integrate it within a localization framework, it is essential to estimate its uncertainty.
ICP errors stem from various error sources.
Among those, the structure of the scene plays an important role: underconstrained situations, such as corridors or mostly flat environments, yield larger errors than geometrically more complex scenes (Fig. \ref{fig:first_page}).

Building on the advances in deep learning techniques dealing with point clouds, we present an approach that estimates data-dependent error models for the ICP process in form of a covariance matrix.
Our method learns he\-te\-ro\-sce\-das\-tic aleatoric uncertainty from ICP input data and uses Bayesian posterior approximation to capture the epistemic uncertainty.

The following section introduces the ICP process, reviews the various sources of errors and works relevant to ours: methods that estimate ICP covariance, data-driven approaches that have proven successful to estimate covariances for other estimation processes, and point-cloud based deep learning techniques. 
Section \ref{sec:learning} formalizes the problem and depicts the learning process. Section \ref{sec:eval} evaluates our results against the state-of-the-art, showing an accurate correlation between the real error of ICP and our predicted covariance.

\section{Background and related work}
\label{sec|background}

\subsection{Iterative Closest Point process}
\label{subsec:icp}

ICP tackles the problem of aligning a \emph{reading} point cloud  $\mathbfcal{P} \in \mathbb{R}^{3 \times n} $ to a \emph{reference} point cloud $\mathbfcal{Q} \in \mathbb{R}^{3 \times m}$, estimating the true rigid transformation between them.
Using an initial estimation $\check{\mT} \in$ SE(3) of the transformation (\eg as provided by a prior on the motion), the rigid transformation $\hat{\mT} \in$ SE(3) that estimates the true transformation is the one that minimizes the error function $e$:

\begin{equation}
\label{eq:icp_min}
\hat{\mT} = \argmin_{\mT}(e( \mT(\mathbfcal{P}), \mathbfcal{Q}))
\end{equation}

where $\mT( \mathbfcal{C} )$ is the transformation $\mT$ applied to the point cloud $\mathbfcal{C}$.
The error function is the result of a weighted ave\-ra\-ge distance between appropriately matched point pairs or point to plane associations.
The optimization is an iterative process, and there is no guarantee to reach the optimal solution.


\subsection{Sources of error for ICP}
\label{subsec:errors}
There are five main sources of error for ICP described in the literature: wrong convergence, wrong initialization, sensor noise and bias, underconstrained situations, and intrinsic randomness of the process \cite{censi2007accurate, brossard2020new}.

\paragraph{Convergence and initialization} Wrong convergence comes from the iterative nature of ICP, which does not guarantee global convergence and can get stuck in local minima.
This is closely related to wrong initialization, which can be a predominant source of error when the initial guess is far from the true solution \cite{brossard2020new}.

\paragraph{Sensor noise and biases} Each sensor naturally suffers changes in its accuracy that depend on several environmental conditions (\eg temperature) \cite{pomerleau2012noise}. 
Also, sensor calibration can introduce distortion \cite{deschaud2018imls}.
While these phenomena introduce a bias in the estimate \cite{laconte2019lidar}, there is an additional source of sensor noise added to each point independently, which generally is a function of the distance, orientation and physical nature or texture of the perceived material.

\paragraph{Underconstrained situations} These occur when the environment does not offer enough information to fully estimate the transformation between the point clouds, mainly by precluding the establishment of good matches between points. 
While in the planar 2D case such situations can be exhaustively identified, namely the corridor and the circle, in the 3D world this becomes a more challenging task. 

\paragraph{Intrinsic ICP randomness} Depending on the con\-fi\-gu\-ra\-tion and implementation, a certain degree of randomness is introduced by subprocesses of ICP, albeit the process itself is deterministic.
For example, filtering clouds with outlier rejection and sub-sampling can yield different estimates, even with the same input data.
\hfill

\subsection{Estimating ICP covariance}
\label{subsec:icp_sota}

Various methods for estimating ICP covariance have been proposed.
Monte-Carlo algorithms \cite{buch2017prediction} generate the covariance from a large number of samples of scan registration results.
Different clouds and initializations are randomly subsampled to obtain a dispersion of the transformation that is used to compute the covariance.
Closed-form methods \cite{censi2007accurate, brossard2020new, prakhya2015closed, gelfand2003geometrically} provide algebraic solutions to the ICP covariance problem by linearizing the error functions defined in the ICP algorithm.
These methods are potentially ill-founded if point reassociations happen at a scale smaller than the actual error \cite{bonnabel2016covariance}.
Experiments on synthetic data have shown that this is often the case \cite{landry2019cello}.
Early closed-form works, as the one in \cite{censi2007accurate}, have been demonstrated over optimistic \cite{mendes2016icp}.
Recently, \cite{landry2019cello} proposed a learning approach to the ICP covariance estimation problem.
The work estimates a 3D covariance based on preprocessed feature descriptors that are used in the learning scheme.
While such descriptors are generic, it is desirable to have a more flexible representation of features impacting on the error of ICP.

It is worth to notice that most of these work tackle dif\-fe\-rent sources of error.
The method presented in \cite{brossard2020new} shows the need for a proper initialization of ICP and proposes a novel approach to 3D uncertainty of ICP accounting for initialization errors.
Censi's pioneering work \cite{censi2007accurate} studies the error due to sensor noise and computes the covariance of this error as a function of the error metric minimized by any ICP implementation.
The work in \cite{landry2019cello}, closer to our approach, explores a data-driven approach to estimate the covariance focusing on sensor noise and aiming at detecting underconstrained situations.

\subsection{Learning motion estimation error models}
\label{subsec:dl_sota}
The literature in deep learning features a large amount of work in the estimation of uncertainty.
The approaches are often tied to the process for which they predict error models.
Interesting works in domains similar to ours have been proposed, mainly tackling the Visual Odometry (VO) case.
As ICP, VO is a well-known family of techniques which estimate a rigid transformation between two consecutive frames for the monocular or stereovision cases.
End-to-end methods perform frame to frame motion estimation \cite{wang2017deepvo} and covariance prediction \cite{wang2018end}.
Both estimates are in this case delegated to neural networks, and do not rely on any geometrical pipeline.
The works in \cite{liu2018deep, demaio2020simultaneously} respectively show covariance and joint bias-covariance estimation to the geometric implementation of VO.

\subsection{Deep learning for point clouds}
\label{subsec:pc_sota}
Recently, deep learning methods applied on point clouds have proven successful for several tasks \cite{qi2017pointnet, qi2017pointnet++, thomas2019kpconv, qi2019deep}.
In particular, PointNet \cite{qi2017pointnet} paved the way for neural networks that directly use 3D point clouds for object classification and segmentation purposes, and has been extended towards learning point cloud features in metric space \cite{qi2017pointnet++}.
This extension increased the effectiveness of PointNet to learn features on different scales, in the manner of traditional convolutional networks.
Other approaches rely on the spherical projection of point clouds, usually coined as range (or depth) image -- which is, in most cases, the original structure of the sensor data.
This is used in the LiDAR odo\-me\-try techniques \cite{cho2019deeplo, wang2020dmlo, li2019net} which makes use of the entire scan to estimate the motion between two frames in an end-to-end manner.

\section{Data driven learning of ICP uncertainty}
\label{sec:learning}

Excluding the process initialization, all error sources of ICP are inherited from the input data.
Sensor noise is directly reflected in the data, as well as ICP subprocesses modifying its size, density and structure (\ie randomness due to filtering).
Additionally, the scene structure, with its inherent geometry and features, largely impacts the error of ICP (as illustrated in Fig. \ref{fig:first_page}), as it defines the observability of the motion along the different dimensions.

Addressing all these sources in a unified framework is a challenging task.
Leveraging recent progresses in deep lear\-ning for point clouds, we address the problem of estimating the covariance of ICP that stems from the last three error sources presented in Sec. \ref{subsec:errors}.
Our hypothesis is that a learning scheme that processes the raw input data can capture these error sources and properly estimate ICP covariances.

In this section we describe our approach to estimate ICP covariances and discuss the nature of the learned uncertainties.
The optimization problem is contextualized in a probabilistic framework and a description of the learning architecture incorporating different types of uncertainty is provided.

\subsection{Minimization problem}
\label{subsec:minim}
Modeling errors for an estimation process can be tackled via a supervised approach.
Consider an estimator, ICP in our case, using sensory data $\mathbfcal{C}_i = \{ \mathbfcal{P}_i, \mathbfcal{Q}_{i+1} \}$ with $\mathbfcal{P}_i \in \mathbb{R}^{n_i}$ and $ \mathbfcal{Q}_{i+1} \in \mathbb{R}^{n_{i+1}}$.
Its output $^i\hat{\mT}_{i+1}$ is an estimate of the ground truth transformation $^i\mT_{i+1}$.
In practice, the point clouds used in the algorithm are the result of several filters \cite{pomerleau2015review}. 
The actual input data is a pair of decimated point clouds $\bar{\mathbfcal{C}}_i = \{ \bar{\mathbfcal{P}}_i, \bar{\mathbfcal{Q}}_{i+1} \}$  with $ \bar{\mathbfcal{P}}_i \in \mathbb{R}^{m_i}, m_i\ll n_i$ and $ \bar{\mathbfcal{Q}}_{i+1} \in \mathbb{R}^{m_{i+1}}, m_{i+1}\ll n_{i+1}$.
The error of ICP is computed as
\begin{equation}
\label{eq:error_dataset}
\ve_i\ =\  ^i\hat{\mT}^{-1}_{i+1}\ \cdot\ ^i\mT_{i+1}
\end{equation}

We represent an error $\ve \in \mathrm{SE(3)}$ as a vector $\vxi \in \mathfrak{se}(3)$ member of the Lie algebra \cite{sola2018micro, barfoot2014associating}
\begin{equation}
\label{eq:xi_se3}
\vxi^{\wedge} = 
\begin{bmatrix}
\vrho \\
\vphi
\end{bmatrix}^{\wedge}
=\ 
\begin{bmatrix}
\vphi^{\wedge} & \vrho\\
\mathbf{0}^\mathsf{T} & 0
\end{bmatrix}
\end{equation}
where the \emph{hat} operator $\wedge$ turns the vector $\vxi \in \mathbb{R}^6$ into a matrix $\vxi^{\wedge} \in \mathbb{R}^{4 \times 4}$.
The opposite linear map is represented as the \emph{vee} operator $\vee$.
The exponential map $\mathrm{exp}(\vxi^{\wedge})$ allows to retract an element of the Lie algebra back to the group. Its inverse is the logarithmic map $\mathrm{log}(\mT)$.
Once the error is defined as $\vxi = \mathrm{log}(\ve)^\vee$, the associated uncertainty $\mSigma \in \mathbb{R}^{6 \times 6}$ can be expressed as
\begin{equation}
\label{eq:cov_se3}
\mSigma = 
\begin{bmatrix}
\mSigma_{\vrho\vrho} & \mSigma_{\vrho\vphi} \\
\mSigma_{\vphi\vrho} & \mSigma_{\vphi\vphi}
\end{bmatrix}
\end{equation}
This uncertainty representation is generally valid for small perturbations that are added to a given pose $\hat{\mT} = \mT \mathrm{exp}(\vxi)$ where $\vxi \sim \mathcal{N}(\mathbf{0}, \mSigma)$\cite{barfoot2014associating}.
The equation above matches the error definition in Eq. \ref{eq:error_dataset}, that is $\vxi \in \mathbb{R}^6$ can be viewed as the error between the real transformation and its estimate.

We define a dataset $\mathcal{D} = \{{\bar{\mathbfcal{C}}}_i, \vxi_i | \forall i \in [1,d] \}$, where $d$ is the size of the dataset.
The goal is to predict the uncertainty of the error vector $\vxi_i$ along its dimensions.
In the data, we noticed that the unbiased error assumption is valid for the vast majority of ICP estimates and the bias is very small otherwise.
To keep the uncertainty assumption valid, we encode a bias term for the error vector in the distribution that ensures the error remains close to a zero-mean Gaussian.
Therefore, it is possible to estimate the parameters of the Gaussian $\mathcal{N}(\vmu, \mSigma)$ as
\begin{equation}
\label{eq:max_prob}
\argmax_{\vmu_{1:d}, \mSigma_{1:d}}\sum_{i=1}^{d} p(\vxi_i | \vmu_i, \mSigma_i)
\end{equation}
To maximize the probability of drawing the correct error given the estimated Gaussian parameters is equivalent to minimize the negative full log-likelihood function \cite{demaio2020simultaneously}.
To ensure a positive definite covariance matrix we use the LDL decomposition as in \cite{liu2018deep}.

\subsection{Uncertainty estimation}
ICP covariance can be decomposed into two categories: epistemic and aleatoric \cite{kendall2017uncertainties}.
Both uncertainties are formulated as probability distributions over different entities.

Aleatoric uncertainty models the uncertainty over the input data to the process.
It assumes an observation noise that can be either constant or vary with the input.
As discussed in Sec.\ref{subsec:errors}, we observed that noise levels change with the scene (Fig. \ref{fig:first_page}), hence we are interested in modeling the va\-ria\-tion of the noise in correlation with the input data. 
This is referred to as heteroscedastic aleatoric uncertainty (as opposed to homoscedastic uncertainty for the constant noise case).

Additionally, it is possible to model the variance intrinsic to the learned model, that is how certain of a given prediction the network is.
Epistemic uncertainty can, in principle, be reduced if more data is used for training. 
In our case, it accounts for the network's poor predictions on out-of-data samples such as scene features that were rarely encountered during training.
Epistemic uncertainty can be modeled by placing a distribution over the network's weights.
Unfortunately, given a prior distribution $\mathcal{N}_{\mW}(0, \sigma)$ over the weights, it is computationally difficult to analytically infer the posterior $p(\mW | \mX, \mY)$, that is the probability of drawing a set of weights given the network input and output.
There is a number of proposed methods to approximate this posterior \cite{stephan2017stochastic, postels2019sampling}.
The work in \cite{gal2016dropout} produces an approximation using dropout variational inference.
This has proven to be an effective yet easily obtainable Bayesian approximation when dealing with deep complex models.

In our context, From Eq.\ref{eq:max_prob}, we consider the mean as a noisy error estimate, and the covariance matrix as the aleatoric uncertainty due to the observation noise.
To approximate the posterior for a Bayesian neural network estimating the ICP errors we resort to Monte-Carlo dropout.
The epistemic uncertainty can be captured by keeping dropout activated at test time.
Sampling $N$ times the predicted ICP error vector and ICP covariance yields a distribution $[\hat{\vmu}_{1:N}, \hat{\mSigma}_{1:N} ] = f_{\mW}(\bar{\mathbfcal{C}})$ where $f_{\mW}(\bar{\mathbfcal{C}})$ is the output of a Bayesian neural network, function of the input point clouds.

The total covariance matrix, integrating aleatoric and epistemic uncertainty is computed as
\begin{equation}
\label{eq:integrated_cov}
\frac{1}{N} \sum_{n=1}^{N} (\hat{\vmu}_n - \bar{\vmu})(\hat{\vmu}_n - \bar{\vmu})^\mathsf{T} + \frac{1}{N} \sum_{n=1}^{N} \hat{\mSigma}_{n}
\end{equation}
where $\bar{\vmu} = \frac{1}{N} \sum_{n=1}^{N} \hat{\vmu}_{n}$ is the mean error vector over the sampled outputs.
The first term in Eq. \ref{eq:integrated_cov} computes the uncertainty over the predicted error vector, which presents a variance proportional to the network confidence over the estimate.
The mean vector is not used to correct ICP estimates as the original error is generally noisy. It is rather evaluated in terms of dispersion after having approximated the model posterior.

\subsection{Learning architecture}
\label{subsec:archi}
The improvements in point clouds based network involved a number of estimation processes.
Recently, FlowNet3D \cite{liu2019flownet3d} proposed an approach to tackle 3D scene flow expanding on the building blocks introduced in \cite{qi2017pointnet++}.
It provides the 3D flow for each point in a reference point cloud deducing its motion from the comparison with a second, successive, point cloud.
Estimating the scene flow is a problem closely related to the scan registration.
Intuitively, a network able to approximate such task results sui\-ta\-ble for ICP related inferences.
In fact, the main difference between the two tasks lies in the generalization needed to shift from a high dimensional flow to the estimate of a rigid transformation in the ICP case.
We leverage the main structure of FlowNet3D to tackle the estimation of ICP errors, modifying few key aspects to adapt it to our problem.
To complete the architecture we add a final MLP layer reducing the dimensionality to the number of parameters needed to predict the Gaussian parameters.
First we follow the original regression layer from FlowNet3D reducing the feature space to $\mathbb{R}^3 \times n$.
The MLP is extended to take as input the feature produced by the regression layer and to return the parameters generating the covariance matrix and predicted error vector.
All the MLP (originally Linear-BatchNorm-ReLU) in FlowNet3D are stacked with Dropout modules to allow the estimation of the epistemic covariance matrix. Additionally, this proved to reduce overfitting as expected.

We initially assumed that including the {\em set upconv} layer would have not been needed.
Intuitively, as in a classical convolutional scheme, the regression task we try to solve can be solved with direct dimensionality reduction, from sensory data to parameters describing uncertainty \cite{demaio2020simultaneously, liu2018deep}.
Nonetheless, retaining upsampling layer produced better results in terms of predicted uncertainty and mean.
We associate this behavior with the architecture using skip connections to use the same features extracted from the same point sets originally used ICP.
This creates a 1:1 matching between the data used by the geometric estimator and the network, leaving less room for outliers.
Our network parameterization, with sampling rates and radius, can be found in Table \ref{table:params}.

\begin{table}
\centering
\small
\begin{tabular}{c c c c}
\rule{0pt}{3ex}Layer type & Radius & Sampling & MLP size\\
\hline
\rule{0pt}{2.5ex}set conv & 2.0 & 1.0$\times$ & [32, 32, 64]\\
\rule{0pt}{2.5ex}set conv & 4.0 & 0.25$\times$ & [64, 64, 128]\\
\rule{0pt}{2.5ex}flow embedding & knn & 1.0$\times$ & [128, 128, 128]\\
\rule{0pt}{2.5ex}set conv & 8.0 & 0.25$\times$ & [128, 128, 256]\\
\rule{0pt}{2.5ex}set conv & 16.0 & 0.125$\times$ & [256, 256, 512]\\
\end{tabular}
\caption{Network architecture parameterization. Set upconv layers are the same as in the updated version of \cite{liu2019flownet3d}.}
\label{table:params}
\end{table}

\section{Evaluation}
\label{sec:eval}
\subsection{ICP process and dataset}
To learn errors produced by a geometric ICP algorithm we selected \emph{libpointmatcher} \cite{pomerleau11tracking}.
It is a widely used open-source library offering a chain that filters point clouds, establishes points associations, removes outliers and estimates rigid transformations using an error minimizer.
For our setup we parameterized the process similarly to \cite{landry2019cello}.
Cloud random sampling is fixed at 2048 random points.
We generate normals, used for the point-to-plane minimizer, using the 10 nearest points before subsampling.

To select the initial transform for ICP, we add an error to the ground truth $\mT$. 
In order not to draw errors from an overoptimistic distribution we sample guesses for the initial transforms from a non-zero mean normal distribution $\mathcal{N}( \mathcal{N}(\mT, a\cdot\mT), b\cdot\mT)$, with $a = 0.25, b = 0.2$.
These figures yield more pessimistic estimates than commonly used aiding sensor (\eg odometry or inertial measurements) posing a challenge to our system.

We select the KITTI odometry dataset \cite{geiger2012ready}, using LiDAR scans, which is made of urban and rural road scenes.
We train and test using all ground truth tagged sequences but sequence \texttt{01}, on which ICP fails to register most of the scans due to scenes lacking features.
To compare against the state of the art in covariance estimation for ICP, we also present results on the challenging data sets of \cite{pomerleau2012challenging}. 
Since for the LiDAR odometry case this dataset is rather small, we refine our KITTI trained model with a few epochs of training, approximately 30 or less for early convergence, on all but one outdoor sequences and validate on the one left out.
As expected, features learned on KITTI are ineffective to predict covariances on the \emph{apartment} and \emph{stairs} sequences, which are highly structured small indoor environments.
Our system learns to model ICP errors with numerically small epistemic uncertainty. 
For reasons of space, we only show results containing the combined aleatoric+epistemic covariance when moving to unseen datasets, \ie applying models learned on KITTI to the ETH dataset.

\subsection{Single-pair validation}
\label{subsec:sp_val}

We first assess the quality of the predicted covariances in a single-pair fashion, and then check their consistence on a full trajectory.
The metrics we use are based on the relation between the observed errors and the predicted covariance.
This is due to the absence of the true distribution which would allow for more classic figures of merit, such as the Kullback-Leibler divergence.
While it would be possible to obtain a pseudo-true distribution by random sampling over the point cloud, this would affect the estimated covariance as well, because the network prediction is tied to the ICP configuration, and most importantly to the input data used by ICP.

As in \cite{censi2007accurate, brossard2020new}, we compute the \emph{Normalized Norm Error (NNE)} to evaluate the scale of the predicted covariance \wrt the actual error:
\begin{equation}
\label{eq:nne}
\mathrm{NNE} = \frac{1}{N}\sum_{n=1}^{N}\sqrt{\frac{||\vxi_n||_2^2}{tr(\hat{\mSigma}_n)}}
\end{equation}
where $tr(\hat{\mSigma}_n)$ is the trace of the predicted covariance matrix.
The optimal value for NNE is one. Values below and above one respectively represent pessimistic and optimistic uncertainty estimates.
Table \ref{table:nne} shows NNE values computed over test KITTI sequences.
The scale of the covariance is well predicted, and is only slightly pessimistic for two validation sequences (\texttt{05} and \texttt{07}).

\begin{table}[h]
\small
\centering
\begin{tabular}{|c||c|c|c|c|c|}
\hline
\rule{0pt}{3ex} \textbf{NNE (KITTI)} & \texttt{05} & \texttt{06} &\texttt{07} &\texttt{09} &\texttt{10}\\
\hline
\rule{0pt}{2.5ex} Translation & 0.70 & 0.61 & 0.58 & 0.94 & 0.90 \\
\Xhline{2\arrayrulewidth}
\rule{0pt}{2.5ex} Rotation & 0.65 & 0.81 & 0.55 & 0.90 & 1.04 \\
\hline
\end{tabular}
\caption{NNE translation and rotation values for the KITTI dataset.}
\label{table:nne}
\end{table}

The NNE is an immediate measure of the adequacy of the predicted variances, but it does not account for off-diagonal covariances. For this purpose, the Mahalanobis distance between ICP estimates and ground truth $\vxi_n = \mathrm{log}( ^n\hat{\mT}^{-1}_{n+1}\ ^n\mT_{n+1})^\vee$, weighted by the predicted covariance $\mSigma_n$ averaged over each consecutive pair in a sequence, is a better metric:
\begin{equation}
\label{eq:dm}
\mathrm{D_M} = \frac{1}{N}\sum_{n=1}^{N} \sqrt{\frac{\vxi_n^{\mathsf{T}} \mSigma_n^{-1} \vxi_n}{dim(\vxi)}}
\end{equation}
As in the NNE case, the optimal value for a normal distribution is one, with distances below and above one respectively denoting pessimistic and optimistic estimates.

\begin{table}
\small
\centering
\begin{tabular}{|c||c|c|c|c|c|}
\hline
\rule{0pt}{4ex} \makecell{\emph{Single-Pair} \\ \textbf{Mah. Dist (KITTI)}} & \texttt{05} & \texttt{06} &\texttt{07} &\texttt{09} &\texttt{10}\\
\hline
\rule{0pt}{2.5ex} Translation & 0.80 & 0.83 & 0.74 & 0.88 & 1.03 \\
\Xhline{2\arrayrulewidth}
\rule{0pt}{2.5ex} Rotation & 0.77 & 0.63 & 0.70 & 0.94 & 0.96 \\
\hline
\end{tabular}
\caption{Single-pair Mahalanobis distance in terms of translation and rotation for the KITTI dataset.}
\label{table:singlepair_dm_kitti}
\vspace{-1.5em}
\end{table}

\begin{figure}[h]
\centering
\captionsetup{width=\linewidth}
\includegraphics[trim={0.35cm 0 0.35cm 0},clip, width=0.95\linewidth]{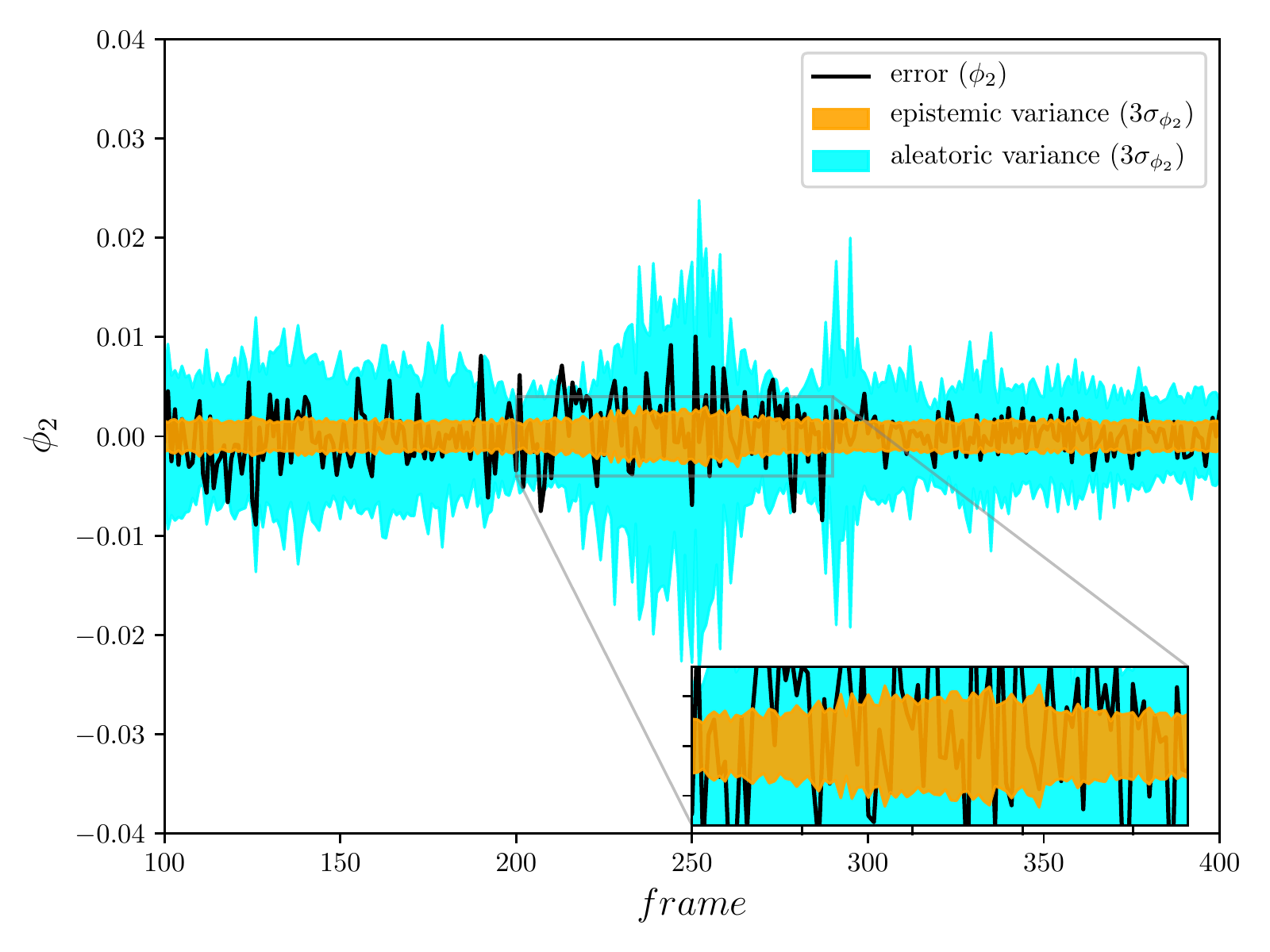}
  \caption{Epistemic and aleatoric $3\sigma$ intervals in a rarely seen area lacking sharp geometrical features.  The real error computed from the ground truth is represented in black. The zoomed-in area shows epistemic uncertainty increase, which corresponds to a {\em wide plain} environment.}
\label{fig:epi_high}
\end{figure}

Table \ref{table:singlepair_dm_kitti} shows the average Mahalanobis distance for the validation sequences.
All sequences present values positively close to 1, indicating that the system is able to correctly capture the uncertainty under the Gaussian assumption.
For these sequences, the epistemic uncertainty is small and constant, which denotes confidence in the system prediction.
Minor variations are encountered for seldom cases: an example is shown in Fig. \ref{fig:epi_high} displaying the epistemic covariance before, during, and after an area corresponding to \emph{wide plain} scenes shown in Fig. \ref{fig:first_page}. Such spatial features are less frequently encountered during training.

\subsection{Trajectory validation}
\label{subsec:traj_val}
In order to assess the quality of our predicted covariances on trajectories we take a multi-step approach.
First, we compose ICP estimates to generate the final pose as $^0\hat{\mT}_f\ =\ ^0\hat{\mT}_1\ ^1\hat{\mT}_2 \cdots\ ^{f-1}\hat{\mT}_f$ for a sequence with $f$ poses.
The ground truth final pose $^0\mT_f$ is computed similarly.
We propagate the covariance associated to each estimate using the 4th order approximation, as in \cite{barfoot2014associating}, to obtain an uncertainty estimate for the final ICP pose.
This way, it is possible to evaluate the Mahalanobis distance (Eq. \ref{eq:dm}) in a spatially consistent manner.
The results are shown in Table \ref{table:traj_dm_kitti}.
They are only slightly optimistic, which is a remarkable outcome given the length of KITTI trajectories (up to more than 2km).

\begin{figure}[]
\centering
\captionsetup{width=\linewidth}
\begin{subfigure}{0.48\columnwidth}
\includegraphics[width=\linewidth]{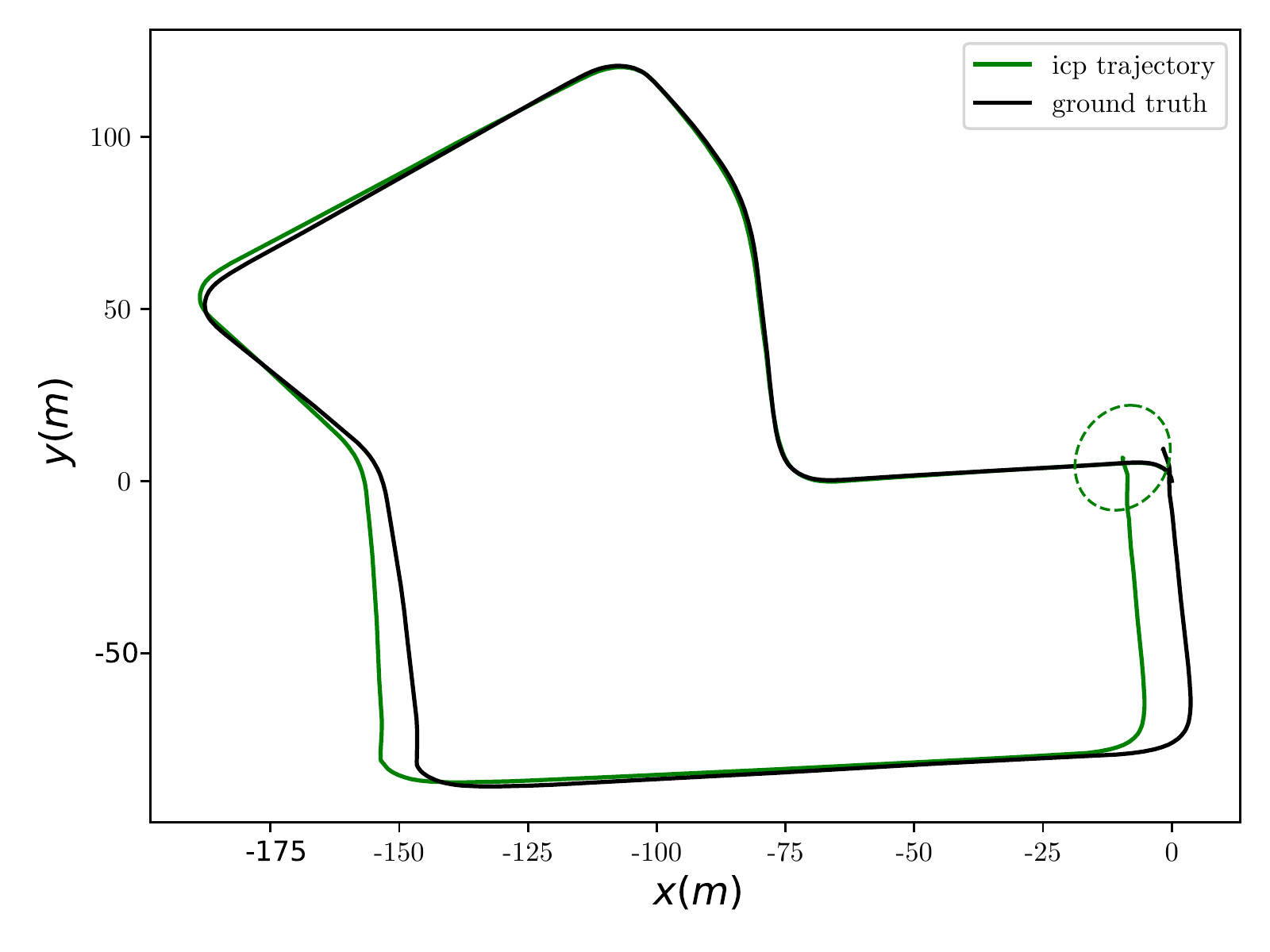}
\end{subfigure}
\begin{subfigure}{0.48\columnwidth}
\includegraphics[width=\linewidth]{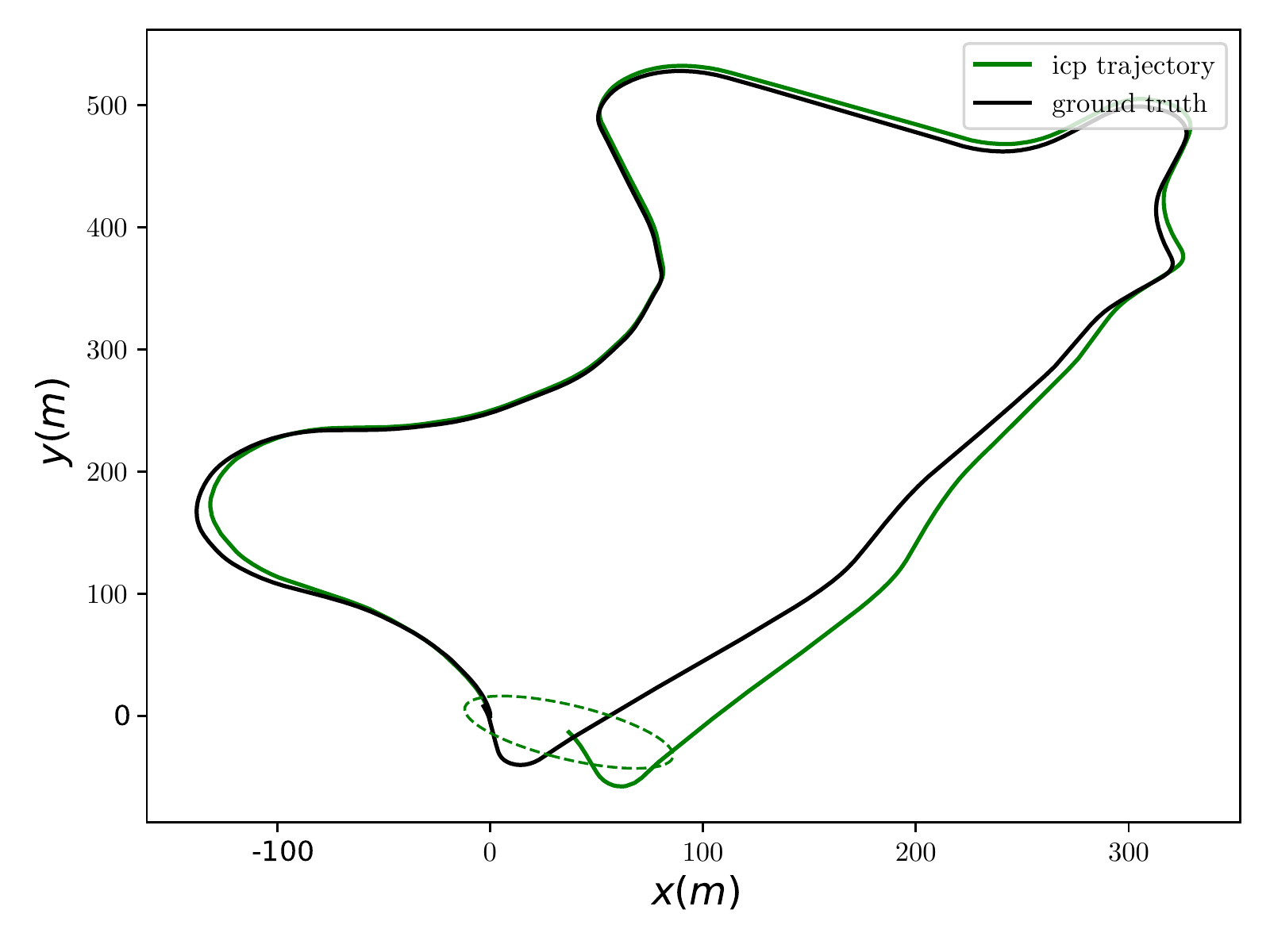}
\end{subfigure}\\
\begin{subfigure}{0.48\columnwidth}
\includegraphics[width=\linewidth]{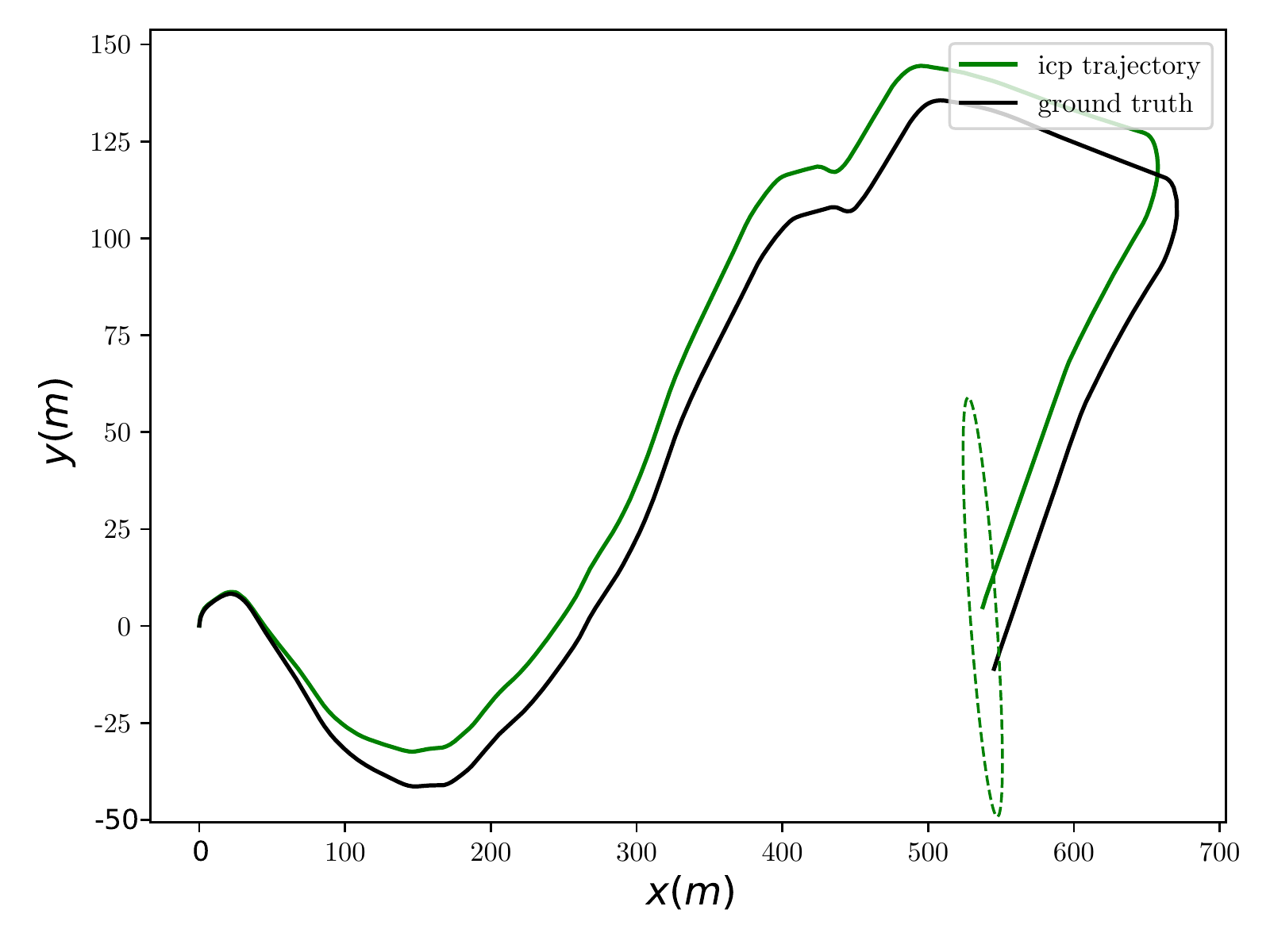}
\end{subfigure}
\begin{subfigure}{0.48\columnwidth}
\includegraphics[width=\linewidth]{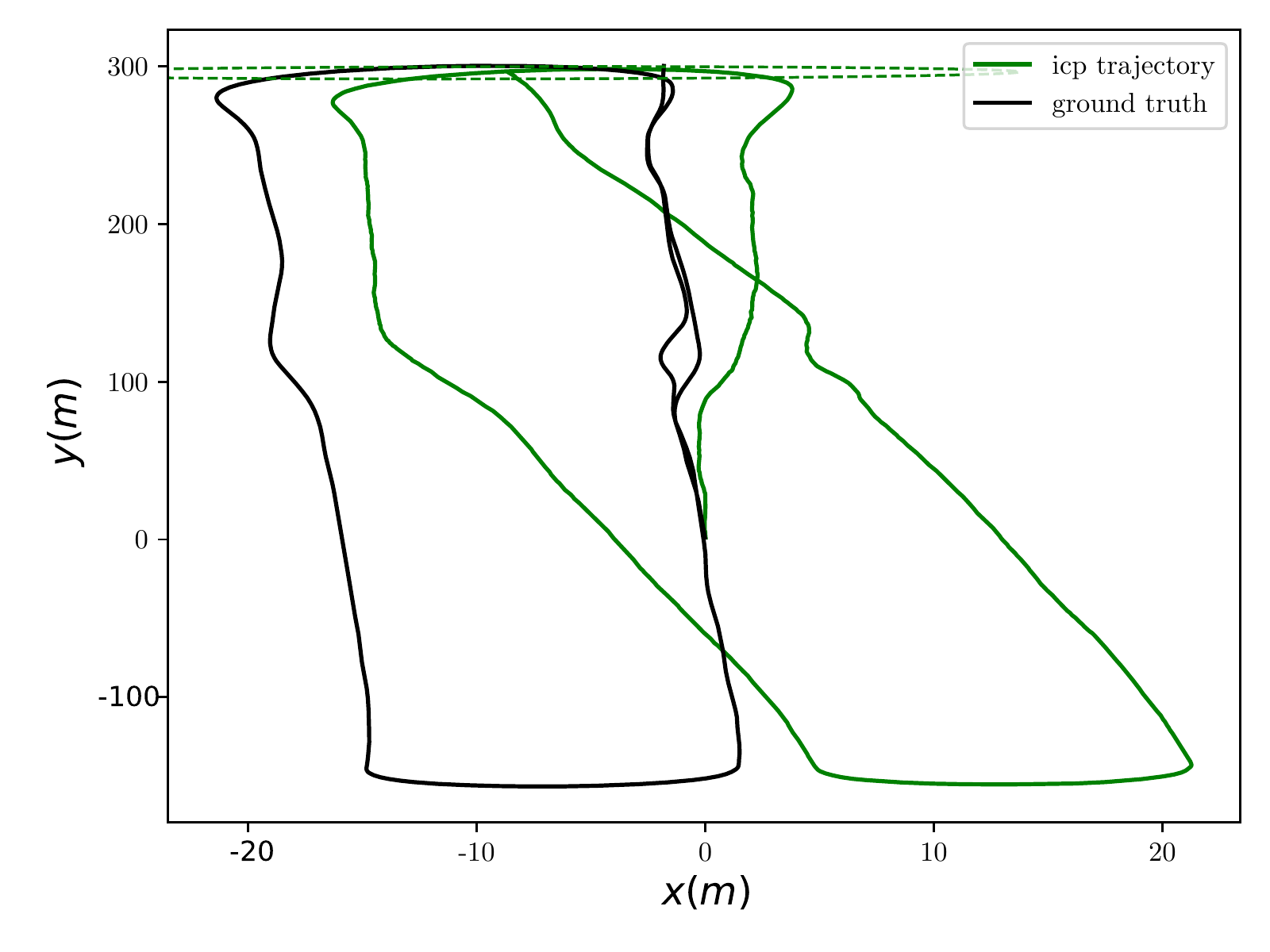}
\end{subfigure}
\caption{Four propagated covariances in the 2$\sigma$ interval (95\%) on considered ICP trajectories compared to ground truth.}
\label{fig:trajectories}
\end{figure}

\begin{table}
\small
\centering
\begin{tabular}{|c||c|c|c|c|c|}
\hline
\rule{0pt}{4ex} \makecell{\emph{Trajectory} \\ \textbf{Mah. Dist (KITTI)}} & \texttt{05} & \texttt{06} &\texttt{07} &\texttt{09} &\texttt{10}\\
\hline
\rule{0pt}{2.5ex} Translation & 2.31 & 1.51 & 1.26 & 1.36 & 4.86 \\
\Xhline{2\arrayrulewidth}
\rule{0pt}{2.5ex} Rotation & 1.49 & 1.95 & 2.19 & 1.89 & 2.95 \\
\hline
\end{tabular}
\caption{Trajectory Mahalanobis distance in terms of translation and rotation for the KITTI dataset.}
\label{table:traj_dm_kitti}
\end{table}

\subsection{Fine tuning on other datasets}
\label{subsec:eth_val}
We also evaluate the results on the smaller dataset of \cite{pomerleau2012challenging} and compare to the state of the art.
As our model requires a significant amount of data we initially learn on a large dataset.
KITTI makes available a considerable number of LiDAR scans with different scenarios, but all correspond to outdoor road environments.
To apply our approach to smaller datasets acquired in very different environments, a fine-tuning process is applied using the pre-trained model on the KITTI dataset.
This allows to obtain much better results compared to re-training on such a small dataset and shows the importance of epistemic uncertainty as an indicator of the performances of a model on unseen data.

\begin{table*}[h]
\small
\centering
\begin{tabular}{|c|c||c|c|c|c|c|c|}
\hline
  \multicolumn{2}{|c||}{\begin{tabular}[c]{@{}c@{}}\textit{Single-Pair}\\ \textbf{Mahalanobis Distance (ETH)}\end{tabular}} & {\begin{tabular}[c]{@{}c@{}}\textit{Gazebo}\\ \textit{winter}\end{tabular}} & {\begin{tabular}[c]{@{}c@{}}\textit{Gazebo}\\ \textit{summer}\end{tabular}} & {\begin{tabular}[c]{@{}c@{}}\textit{Mountain}\\ \end{tabular}} &
              {\begin{tabular}[c]{@{}c@{}}\textit{Haupt-}\\ \textit{-gebaude}\end{tabular}} &
              {\begin{tabular}[c]{@{}c@{}}\textit{Wood}\\ \textit{autumn}\end{tabular}} &
              {\begin{tabular}[c]{@{}c@{}}\textit{Wood}\\ \textit{summer}\end{tabular}} \\
\hline
\rowcolor{Gray}
\rule{0pt}{2.5ex} \multirow{2}{*}{\cellcolor{white}} & Trans (A) & 2.05 & 2.32 & 4.15 & 2.68 & 4.01 & 4.03 \\
\cline{2-8}
\rule{0pt}{2.5ex}\cellcolor{white}\raisebox{1.6ex}[1.6ex]{\textbf{Ours} (Direct)} & Trans (E+A) & 2.04 & 2.12 & 3.88 & 2.49 & 4.19 & 3.78\\
\hline
\rowcolor{Gray}
\rule{0pt}{2.5ex} \multirow{2}{*}{\cellcolor{white}} & Trans (A) & \textbf{0.64} & 1.12 & 1.27 & 1.04 & 1.86 & \textbf{1.03} \\
\cline{2-8}
\rule{0pt}{2.5ex}\cellcolor{white}\raisebox{1.6ex}[1.6ex]{\textbf{Ours} (Finetuned)} & Trans (E+A) & 0.59 & \textbf{1.06} & \textbf{1.21} & \textbf{0.98} & \textbf{1.70} & 0.90\\
\Xhline{3\arrayrulewidth}
\rowcolor{Gray}
\rule{0pt}{2.5ex} \multirow{2}{*}{\cellcolor{white}} & Rot (A) & 1.33 & 1.93 & 3.06 & 2.61 & 2.10 & 2.33 \\
\cline{2-8}
\rule{0pt}{2.5ex} \cellcolor{white}\raisebox{1.6ex}[1.6ex]{\textbf{Ours} (Direct)} & Rot (E+A) & \textbf{1.30} & 1.89 & 3.03 & 2.47 & 2.07 & 2.19 \\
\hline
\rowcolor{Gray}
\rule{0pt}{2.5ex} \multirow{2}{*}{\cellcolor{white}} & Rot (A) & 0.48 & \textbf{0.53} & \textbf{0.65} & \textbf{1.02} & \textbf{0.54} & \textbf{0.82} \\
\cline{2-8}
\rule{0pt}{2.5ex} \cellcolor{white}\raisebox{1.6ex}[1.6ex]{\textbf{Ours} (Finetuned)} & Rot (E+A) & 0.44 & 0.49 & \textbf{0.65} & 0.96 & 0.51 & 0.71 \\
\hline
\end{tabular}
\caption{Single-pair Mahalanobis distance in terms of translation and rotation for the ETH dataset using aleatoric only (A) and combined epistemic-aleatoric (E+A) uncertainty. A comparison between the pre-trained model on KITTI and after finetuning is presented. After finetuning, the epistemic uncertainty is less prominent and its (minor) impact on performances depends on whether the system is being pessimistic or optimistic.}
\label{table:singlepair_dm_eth}
\end{table*}

\begin{table*}[h]
\small
\centering
\begin{tabular}{|c|c||c|c|c|c|c|c|}
\hline
  \multicolumn{2}{|c||}{\begin{tabular}[c]{@{}c@{}}\textit{Trajectory}\\ \textbf{Mahalanobis Distance (ETH)}\end{tabular}} & {\begin{tabular}[c]{@{}c@{}}\textit{Gazebo}\\ \textit{winter}\end{tabular}} & {\begin{tabular}[c]{@{}c@{}}\textit{Gazebo}\\ \textit{summer}\end{tabular}} & {\begin{tabular}[c]{@{}c@{}}\textit{Mountain}\\ \end{tabular}} &
              {\begin{tabular}[c]{@{}c@{}}\textit{Haupt-}\\ \textit{-gebaude}\end{tabular}} &
              {\begin{tabular}[c]{@{}c@{}}\textit{Wood}\\ \textit{autumn}\end{tabular}} &
              {\begin{tabular}[c]{@{}c@{}}\textit{Wood}\\ \textit{summer}\end{tabular}} \\
  \hline
\rowcolor{white}
\rule{0pt}{2.5ex} \cellcolor{white} CELLO-3D \cite{landry2019cello}& Trans & 0.1 & 0.2 & - & 0.3 & 0.1 & 0.1 \\
\hline
\rule{0pt}{2.5ex} \emph{Brossard et al.} \cite{brossard2020new} & Trans & 1.8 & \textbf{1.0} & 1.2 & \textbf{1.8} & 1.2 & 1.5 \\
\hline
\rowcolor{white}
\hline
\rowcolor{white}
\rule{0pt}{2.5ex} \cellcolor{white}\textbf{Ours} (Finetuned) & Trans (A) & \textbf{0.83} & 1.28 & \textbf{0.87} & 1.83 & \textbf{0.93} & \textbf{0.69} \\
\cline{2-8}

\Xhline{3\arrayrulewidth}
\rowcolor{white}
\rule{0pt}{2.5ex} \cellcolor{white} CELLO-3D \cite{landry2019cello}& Rot & 0.2 & 0.2 & - & 0.2 & 0.3 & 0.3 \\
\hline
\rowcolor{white}
\rule{0pt}{2.5ex} \cellcolor{white}\emph{Brossard et al.} \cite{brossard2020new} & Rot & 3.7 & 2.3 & 1.2 & 2.9 & 4.2 & 4.7 \\
\hline
\rowcolor{white}
\rule{0pt}{2.5ex} \textbf{Ours} (Finetuned) & Rot (A) & \textbf{0.71} & \textbf{1.03} & \textbf{0.81} & \textbf{1.48} & \textbf{1.69} & \textbf{1.65} \\
\hline
\end{tabular}
\caption{Trajectory Mahalanobis distance in terms of translation and rotation for the ETH dataset using aleatoric only (A) uncertainty, compared to two state-of-the-art approaches. Epistemic uncertainty is not showed after finetuning.}
\label{table:traj_dm_eth}
\vspace{-1em}
\end{table*}

Table \ref{table:singlepair_dm_eth} shows the result of our approach on single-pair estimates with the dataset of \cite{pomerleau2012challenging}.
While the environments are very different, the model performs remarkably well in outdoor scenes.
Even architectural features, such as in \emph{hauptgebaude} can be leveraged to produce faithful covariance predictions as, intuitively, the scene presents a structure (flat ground plane, lack of indoor man-made elements) that is geometrically similar to the some scenes encountered in KITTI.

\begin{figure}[h]
\centering
\captionsetup{width=\linewidth}
\includegraphics[trim={0.25cm 0 0.35cm 0},clip,width=0.87\linewidth]{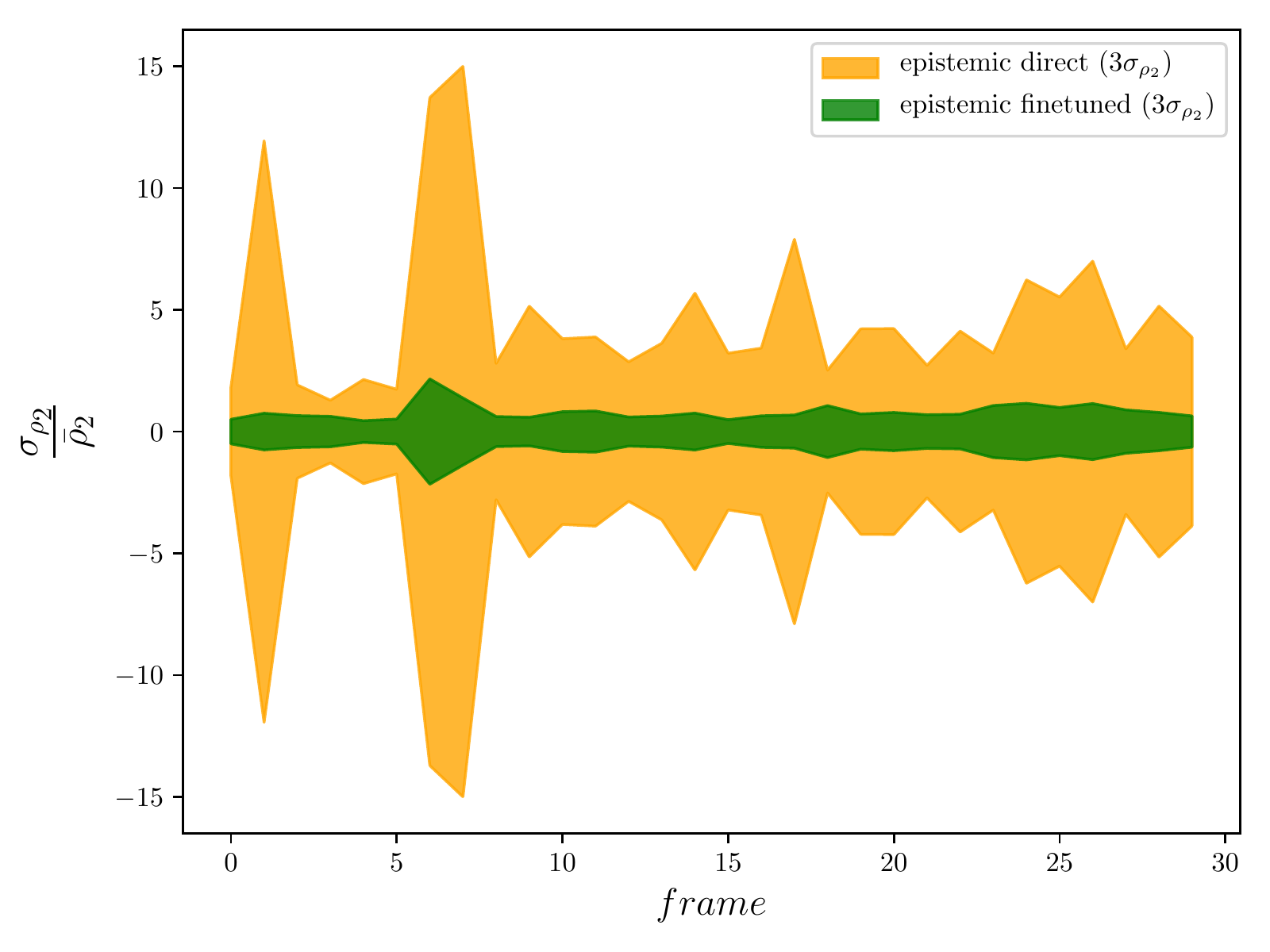}
  \caption{Epistemic uncertainties normalized by the mean in the direct and finetuned case.}
\label{fig:direct_vs_finetuned}
\end{figure}

More importantly, Table \ref{table:singlepair_dm_eth} highlights the impact of the finetuning, not only in the expected improved results, but also in the magnitude of the epistemic uncertainty.
The improvement brought by considering the model uncertainty is evident when testing the model trained on KITTI directly on an unseen dataset, where, as expected, the network has little confidence in its prediction for new scenarios.
In fact, while some outdoor features may recur in the ETH dataset, motion type and scene size are remarkably different when compared to the KITTI dataset.
After finetuning, not only the weight of the epistemic covariance is greatly reduced as shown in Fig. \ref{fig:direct_vs_finetuned}, but the overall results are closer to the optimal value.


Finally, Table \ref{table:traj_dm_eth} compares our results against CELLO-3D \cite{landry2019cello} and \textit{Brossard et al.} \cite{brossard2020new}  on aggregated transforms on the same dataset.
It is worth noticing that given the strong focus on ICP initialization from \textit{Brossard et al.}, it is more relevant to compare our approach to CELLO-3D, 
as our work solely focus on the errors introduced directly and indirectly by the data on a purely data-driven approach.
CELLO-3D is often over pessimistic while \emph{Brossard et al.} is generally optimistic.
Especially on rotational uncertainty, our method outperforms the two approaches by a significant margin.

\section{Conclusion}
\label{sec:conclusions}
We presented an approach to estimate faithful covariances for the ICP scan registration process.
Using a data-driven paradigm we have shown that it is possible to learn uncertainty for the selected algorithm, and to generalize to a certain degree on multiple datasets.
Leveraging a neural network architecture conceived to directly process point clouds, we estimate covariances with a close connection to the data used in the ICP algorithm.
Standard metrics show the reliability of our approach, and that it generally outperforms the state-of-the-art.
Future improvements will concern the network architecture, possibly tailoring it to the scan registration problem.
A possible approach is to have the flow embedding layer (and the successive layers) doubled to account for both point clouds motions, whereas in the scene flow the problem is cast on the reference cloud only.
Additionally, the role of initialization and the use of the initial guess in the learning process can also be considered.

\newpage
{\small
\bibliographystyle{ieeetr}
\bibliography{phdlaas}
}

\end{document}

%% file: math_commands.tex
\usepackage{amsmath,amsfonts,bm}

\def\eqref#1{equation~\ref{#1}}

\def\1{\bm{1}}

\def\vmu{{\bm{\mu}}}

\def\ve{{\bm{e}}}

\def\vxi{{\bm{\xi}}}
\def\vrho{{\bm{\rho}}}
\def\vphi{{\bm{\phi}}}

\def\mT{{\bm{T}}}

\def\mW{{\bm{W}}}
\def\mX{{\bm{X}}}
\def\mY{{\bm{Y}}}

\def\mSigma{{\bm{\Sigma}}}

\DeclareMathAlphabet{\mathsfit}{\encodingdefault}{\sfdefault}{m}{sl}
\SetMathAlphabet{\mathsfit}{bold}{\encodingdefault}{\sfdefault}{bx}{n}

\DeclareMathOperator*{\argmax}{arg\,max}
\DeclareMathOperator*{\argmin}{arg\,min}

\DeclareMathAlphabet\mathbfcal{OMS}{cmsy}{b}{n}